\documentclass[11pt,a4paper]{article}
\usepackage[hyperref]{acl2019}
\usepackage{times}
 \usepackage{xcolor}
\usepackage{latexsym}
\usepackage{amssymb}
\usepackage{amsmath}
\usepackage{array}

\usepackage{booktabs} 
\usepackage{graphicx}
\usepackage{setspace}

\usepackage{subfigure}
\usepackage{url}
\aclfinalcopy 

\title{BiSET: Bi-directional Selective Encoding with Template for Abstractive Summarization}

\author{Kai Wang \\
  Sun Yat-sen University \\
  \texttt{\small wangk73@mail2.sysu.edu.cn} \\\And
  Xiaojun Quan\thanks{ \hspace*{0.15cm}Corresponding author.} \\
  Sun Yat-sen University \\
  \texttt{\small quanxj3@mail.sysu.edu.cn} \\\And
  Rui Wang \\
  Alibaba Inc.\\
  \texttt{\small masi.wr@alibaba-inc.com} \\}

\date{}

\begin{document}
\maketitle
\begin{abstract}
The success of neural summarization models stems from the meticulous encodings of source articles. To overcome the impediments of limited and sometimes noisy training data, one promising direction is to make better use of the available training data by applying filters during summarization. In this paper, we propose a novel \textbf{Bi}-directional \textbf{S}elective \textbf{E}ncoding with \textbf{T}emplate (BiSET) model, which leverages template discovered from training data to softly select key information from each source article to guide its summarization process. Extensive experiments on a standard summarization dataset were conducted and the results show that the template-equipped BiSET model manages to improve the summarization performance significantly with a new state of the art.
\end{abstract}

\section{Introduction}
Abstractive summarization aims to shorten a source article or paragraph by rewriting while preserving the main idea. Due to the difficulties in rewriting long documents, a large body of research on this topic has focused on paragraph-level article summarization. Among them, sequence-to-sequence models have become the mainstream and some have achieved state-of-the-art performance \cite{Rush2015A,Chopra2016Abstractive,Nallapati2016Abstractive}. In general, the only available information for these models during decoding is simply the source article representations from the encoder and the generated words from the previous time steps \cite{Nallapati2016Abstractive,Gu2016Incorporating,Lin2018Global}, while the previous words are also generated based on the article representations. Since natural language text is complicated and verbose in nature, and training data is insufficient in size to help the models distinguish important article information from noise, sequence-to-sequence models tend to deteriorate with the accumulation of word generation, e.g., they generate irrelevant and repeated words frequently \cite{koehn2017six}.

Template-based summarization \cite{zhou2004template} is an effective approach to traditional abstractive summarization, in which a number of hard templates are manually created by domain experts, and key snippets are then extracted and populated into the templates to form the final summaries. The advantage of such approach is it can guarantee concise and coherent summaries in no need of any training data. However, it is unrealistic to create all the templates manually since this work requires considerable domain knowledge and is also labor-intensive. Fortunately, the summaries of some specific training articles can provide similar guidance to the summarization as hard templates. Accordingly, these summaries are referred to as soft templates, or templates for simplicity, in this paper.

Despite their potential in relieving the verbosity and insufficiency problems of natural language data, templates have not been exploited to full advantage. For example, \newcite{cao2018retrieve} simply concatenated template encoding after the source article in their summarization work. To this end, we propose a Bi-directional Selective Encoding with Template (BiSET) model for abstractive sentence summarization. Our model involves a novel bi-directional selective layer with two gates to mutually select key information from an article and its template to assist with summary generation. Due to the limitations in obtaining handcrafted templates, we further propose a multi-stage process for automatic retrieval of high-quality templates from training corpus. Extensive experiments were conducted on the Gigaword dataset \cite{Rush2015A}, a public dataset widely used for abstractive sentence summarization, and the results appear to be quite promising. Merely using the templates selected by our approach as the final summaries, our model can already achieve superior performance to some baseline models, demonstrating the effect of our templates. This may also indicate the availability of many quality templates in the corpus. Secondly, the template-equipped summarization model, BiSET, outperforms all the state-of-the-art models significantly. To evaluate the importance of the bi-directional selective layer and the two gates, we conducted an ablation study by discarding them respectively, and the results show that, while both of the gates are necessary, the template-to-article (T2A) gate tends to be more important than the article-to-template (A2T) gate. A human evaluation further validates the effectiveness of our model in generating informative, concise and readable summaries.

\begin{spacing}{1.0}
The contributions of this work include:
\begin{itemize}
  \setlength\itemsep{0.01em}
  \item We propose a novel bi-directional selective mechanism with two gates to mutually select important information from both article and template to assist with summary generation.
  \item We develop a Fast Rerank method to automatically select high-quality templates from training corpus.
  \item Empirical evaluations on the benchmark dataset show our model has achieved a new state of the art.
  \item The source code of this work has been released for future research.\footnote{https://github.com/InitialBug/BiSET}
\end{itemize}

\section{The Framework}
Our framework includes three key modules: Retrieve, Fast Rerank, and BiSET. For each source article, Retrieve aims to return a few candidate templates from the training corpus. Then, the Fast Rerank module quickly identifies a best template from the candidates. Finally, BiSET mutually selects important information from the source article and the template to generate an enhanced article representation for summarization.

\begin{figure*}

\centering
\includegraphics[width=5in]{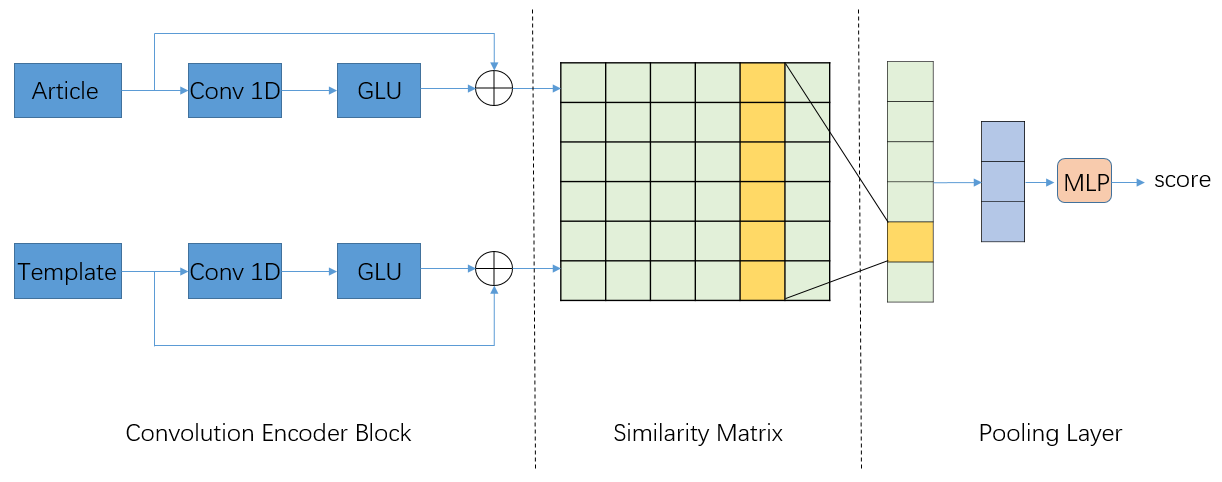}
\caption{Overview of the Fast Rerank Module.}
\label{img:rerank}
\end{figure*}

\subsection{Retrieve}
This module starts with a standard information retrieval library\footnote{https://lucene.apache.org} to retrieve a small set of candidates for fine-grained filtering as \newcite{cao2018retrieve}. To do that, all non-alphabetic characters (e.g., dates) are removed to eliminate their influence on article matching. The retrieval process starts by querying the training corpus with a source article to find a few (5 to 30) related articles, the summaries of which will be treated as candidate templates.

\subsection{Fast Rerank}\label{sec:rerank}
The above retrieval process is essentially based on superficial word matching and cannot measure the deep semantic relationship between two articles. Therefore, the Fast Rerank module is developed to identify a best template from the candidates based on their deep semantic relevance with the source article. We regard the candidate with highest relevance as the template. As illustrated in Figure \ref{img:rerank}, this module consists of a Convolution Encoder Block, a Similarity Matrix and a Pooling Layer.

\noindent\textbf{Convolution Encoder Block}. This block maps the input article and its candidate templates into high-level representations. The popular ways to this are either by using recurrent neural network (RNN) or a stack of convolutional neural network (CNN), while none of them are suitable for our problem. This is because a source article is usually much longer than a template, and both RNN and CNN may lead to semantic irrelevance after encodings. Instead, we implement a new convolution encoder block which includes a word embedding layer, a 1-D convolution followed by a non-linearity function, and residual connections \cite{gehring2017convolutional}.

Formally, given word embeddings $\left \{ e _{i} \right \}_{i=1}^{E}\in \mathbb{R}^d$ of an article, we use a 1-D convolution with kernel $\textbf{k}\in \mathbb{R}^{2d\times kd}$ and bias $b_h\in \mathbb{R}^{2d}$ to extract the n-gram features:
\begin{equation}
	h_i=\textbf{k}[e_{i-k/2},...,e_{i+k/2}]+b_h
\end{equation}
where $h_i\in \mathbb{R}^{2d}$. We pad both sides of an article/template with zeros to keep fixed length. After that, we employ the gated linear unit (GLU) \cite{dauphin2017language} as our activation function to control the proportion of information to pass through. GLU takes half the dimension of $h_i$ as input and reduces the input dimension to $d$. Let $h_i=[h_i^1;h_i^2]$, where $h_i^1,h_i^2\in \mathbb{R}^d$, we have:
\begin{equation}
	\resizebox{.8\hsize}{!}{$r_i=GLU(h_i)=GLU([h_i^1;h_i^2])=h_i^1\otimes \sigma(h_i^2)$}
\end{equation}
where $r_i\in \mathbb{R}^d$, $\sigma$ is the sigmoid function, and $\otimes$ means element-wise multiplication. To retain the original information, we add residual connections from the input of the convolution layer to the output of this block: $z_i=r_i+e_i$.

\noindent\textbf{Similarity Matrix}. The above encoder block generates a high-level representation for each source article/candidate template. Then, a similarity matrix $\mathcal{S}\in \mathbb{R}^{m\times n}$ is calculated for a given article representation, $\textbf{S}\in \mathbb{R}^{m\times d}$, and a template representation, $\textbf{T}\in \mathbb{R}^{n\times d}$:
\begin{equation}
	s_{ij}=f(\textbf{S}_i,\textbf{T}_j)
\end{equation}
where $f$ is the similarity function, and the common options for $f$ include:
\begin{equation}
f(x,y)=
\begin{cases}
x^Ty,& \text{dot product}\\
x^TWy,& \text{bilinear function}\\
\|x-y\|,& \text{Euclidean distance}
\end{cases}
\end{equation}
Most previous work uses dot product or bilinear function \cite{chen2016thorough} for the similarity, yet we find the family of Euclidean distance perform much better for our task. Therefore, we define the similarity function as:
\begin{equation}
	f(x,y)=exp(-\|x-y\|^2)
\end{equation}

\noindent\textbf{Pooling Layer}. This layer is intended to filter out unnecessary information in the matrix $\mathcal{S}$. Before applying such pooling operations as max-pooling and k-max pooling \cite{kalchbrenner2014convolutional} over the similarity matrix, we note there are repeated words in the source article, which we only want to count once. For this reason, we first identify some salient weights from $\mathcal{S}$:
\begin{equation}
	q=max_{column}(\mathcal{S})
\end{equation}
where $max_{column}$ is a column-wise maximum function. We then apply k-max pooling over $q$ to select $k$ most important weights, $p\in \mathbb{R}^k$. Finally, we apply a two-layer feed-forward network to output a similarity score for the source article and the candidate template:
\begin{gather}
    p=k\textrm{-}max(q)\\
    a=ReLU(\textbf{W}_ap+b_1)\\
	\label{eq:10} s=\sigma(\textbf{W}_sa+b_2)
\end{gather}

\begin{figure*}[t]
\centering
\subfigure[]{
\label{img:biset}
\includegraphics[scale=0.16]{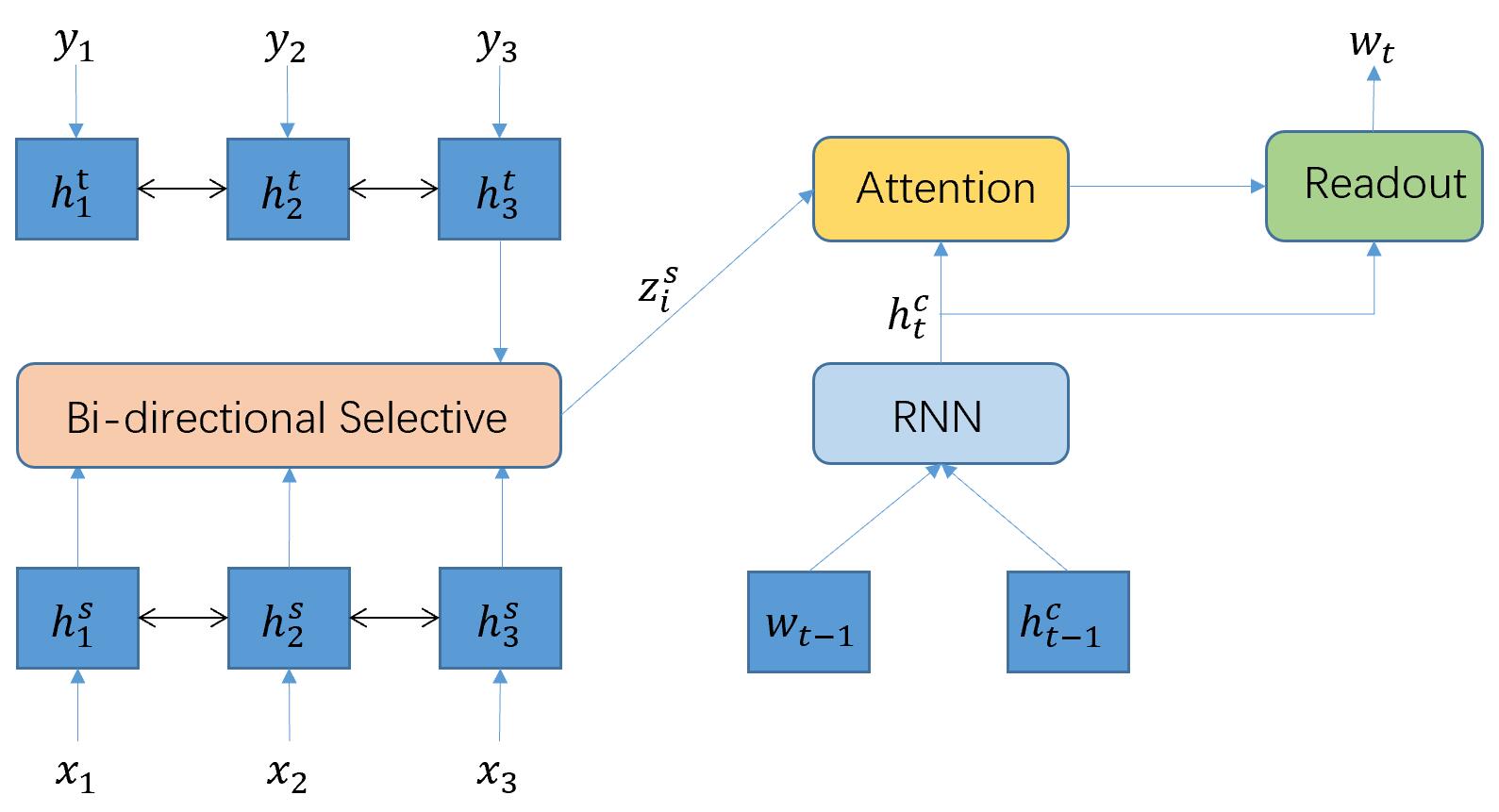}}
\hspace{0.1in}
\subfigure[]{
\label{img:gate} 
\includegraphics[scale=0.16]{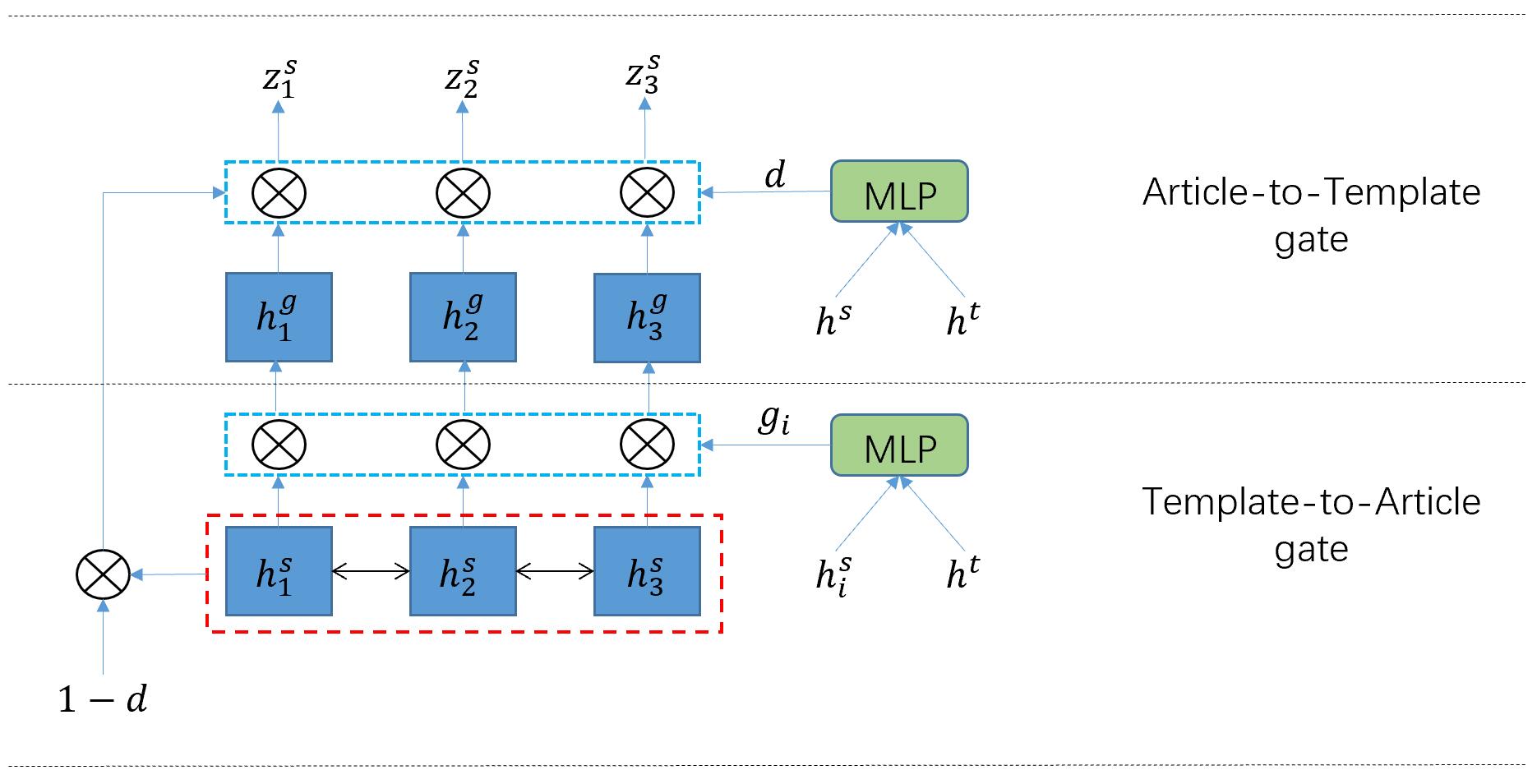}}
\caption{The structure of the proposed model: (a) the Bi-Directional Selective Encoding with Template model (BiSET) and (b) the bi-directional selective layer.}
\label{img:model} 
\end{figure*}


\subsection{Traditional Methodologies}\label{section:interaction_methods}
In this section, we explore three traditional approaches to taking advantage of the templates for summarization. They share the same encoder and decoder layers, but own different interaction layers for combination of a source article and template. The encoder layer uses a standard bi-directional RNN (BiRNN) to separately encode the source article and the template into hidden states $h_i^s$ and $h_j^t$.

\noindent\textbf{Concatenation}. This approach directly concatenates the hidden state, $\left \{ h _{i}^t \right \}_{i=1}^{N}$, of a template after the article representation, $\left \{ h _{i}^s \right \}_{i=1}^{M}$, to form a new article representation, $\left \{ z _{i}^s \right \}_{i=1}^{M+N}$. This approach is similar to $R^3Sum$ \cite{cao2018retrieve} but uses our Fast Rerank and summary generation modules.

\noindent\textbf{Concatenation+Self-Attention}.~This approach adds a multi-head self-attention  \cite{Vaswani2017Attention} layer with 4 heads on the basis of the above direct concatenation.

\noindent\textbf{DCN Attention}. Initially introduced for machine reading comprehension \cite{seo2016bidirectional}, this interaction approach is employed here to create template-aware article representations. First, we compute a similarity matrix, $\mathcal{S}\in \mathbb{R}^{m\times n}$, for each pair of article and template words by $s_{ij}=\textbf{W}_0[h_i^s;h_j^t;h_i^s\otimes h_j^t]$, where `;' is the concatenation operation. We then normalize each row and column of $\mathcal{S}$ by softmax, giving rise to two new matrices $\overline{\mathcal{S}}$ and $\overline{\overline{\mathcal{S}}}$. After that, the Dynamic Coattention Network (DCN) attention is applied to compute the bi-directional attention: $\textbf{A}=\overline{\mathcal{S}}\cdot h^t$ and $\textbf{B}=\overline{\mathcal{S}}\cdot \overline{\overline{\mathcal{S}}}^T\cdot h^s$, where $\textbf{A}$ denotes article-to-template attention and $\textbf{B}$ is template-to-article attention. Finally, we obtain the template-aware article representation $\left \{ z _{i}^s \right \}_{i=1}^{M}$:
\begin{equation}
z_i^s=[h_i^s;\textbf{A}_i;h_i^s\otimes \textbf{A}_i;h_i^s\otimes \textbf{B}_i]
\end{equation}

\subsection{BiSET}
Inspired by the research in machine reading comprehension \cite{seo2016bidirectional} and selective mechanism \cite{Zhou2017Selective}, we propose a novel \textbf{Bi}-directional \textbf{S}elective \textbf{E}ncoding with \textbf{T}emplate (BiSET) model for abstractive sentence summarization. The core idea behind BiSET is to involve templates to assist with article representation and summary generation.~As shown in Figure \ref{img:model}, BiSET contains two selective gates: Template-to-Article (T2A) gate and Article-to-Template (A2T) gate. The role of T2A is to use a template to filter the source article representation:
\begin{gather}
    g_i=\sigma(\textbf{W}_{sh} h_i^s+\textbf{W}_{th} h^t+b_s)\\
    h_i^g=h_i^s\otimes g_i
\end{gather}
where $h^t$ is the concatenation of the last forward hidden state, $\overrightarrow{h_n^t}$, and the first backward hidden state, $\overleftarrow{h_1^t}$, of the template.

On the other hand, the purpose of A2T is to control the proportion of $h^g$ in the final article representation. We assume the source article is credible and use its representation $h^s$ together with $h^t$ to calculate a confidence degree, where $h^s$ is obtained in a similar way as $h^t$. The confidence degree $d$ is computed by:
\begin{equation}
    d=\sigma((h^s)^T\textbf{W}_d h^t+b_d)
\end{equation}
The final source article representation is calculated as the weighted sum of $h_i^s$ and $h_i^g$:
\begin{equation}
    z_i^s=dh_i^g+(1-d)h_i^s
\end{equation}
which allows a flexible manner for template incorporation and helps to resist errors when low-quality templates are given.

\noindent\textbf{The decoder layer}. This layer includes an ordinary RNN decoder \cite{Luong2015Effective}. At each time step $t$, the decoder reads the word $w_{t-1}$ and hidden state $h_{t-1}^c$ generated in the previous step, and gives a new hidden state for the current step:
\begin{gather}
	h_t^c=RNN(w_{t-1},h_{t-1}^c)
\end{gather}
where the hidden state is initialized with the original source article representation, $h^s$. We then compute the attention between $h_t^c$ and the final article representation $z^s$ to obtain a context vector $c_t$:
\begin{gather}
    \varepsilon_{t,i}=(z_i^s)^T\textbf{W}_ch_t^c\\
	\alpha_{t,i}=\frac{exp(\varepsilon_{t,i})}{\sum_{i=1}^{M}exp(\varepsilon_{t,i})}\\
   c_t=\sum\limits_{i=1}^M\alpha_{t,i}z_i^s
\end{gather}
After that, a simple concatenation layer is used to combine the hidden state $h_t^c$ and the context vector $c_t$ into a new hidden state $h_t^a$:
\begin{equation}
   h_t^a=tanh(\textbf{W}_{ha}[c_t;h_t^c])
\end{equation}
which will be mapped to a new representation of vocabulary size and fed through a softmax layer to output the target word distribution:
\begin{equation}
   p(w_t|w_1,...,w_{t-1})=softmax(\textbf{W}_p h_t^a)
\end{equation}

\subsection{Training}
The Retrieve module involves an unsupervised process with traditional indexing and retrieval techniques. For Fast Rerank, since there is no ground truth available, we use ROUGE-1\footnote{We also tried ROUGE-2 and ROUGE-L, but ROUGE-1 shows to be more suitable.} \cite{Lin2003Automatic} to evaluate the saliency of a candidate template with respect to the gold summary of current source article. Therefore, the loss function is defined as:
\begin{equation}
L_r(\theta)=-\frac{1}{N}\sum\limits_{i=1}^N[s^*\log s+(1-s^*)\log(1-s)]
\end{equation}
where $s$ is a score predicted by Equation \ref{eq:10}, and $N$ is the product of the training set size, $D$, and the number of retrieved templates for each article.

For the BiSET module, the loss function is chosen as the negative log-likelihood between the generated summary, $w$, and the true summary, $w^*$:
\begin{equation}
L_w(\theta)=-\frac{1}{D}\sum\limits_{i=1}^D\sum\limits_{j=1}^L\log p(w^{*(i)}_j|w_{j-1}^{(i)},x^{(i)},y^{(i)})
\end{equation}
where $L$ is the length of the true summary, $\theta$ contains all the trainable variables, and $x$ and $y$ denote the source article and the template, respectively.

\section{Experiments}
In this section, we introduce our evaluations on a standard dataset.

\subsection{Dataset and Implementation}
The dataset used for evaluation is Annotated English Gigaword \cite{Napoles2012Annotated}, a parallel corpus formed by pairing the first sentence of an article with its headline. For a fair comparison, we use the version preprocessed by \newcite{Rush2015A}\footnote{ https://github.com/harvardnlp/sent-summary} as previous work.

During training, both the Fast Rerank and BiSET modules have a batch size of 64 with the Adam optimizer \cite{Kingma2014Adam}. We also apply grad clipping \cite{pascanu2013difficulty} with a range of [-5,5]. The differences of the two modules in settings are listed below.

\noindent\textbf{Fast Rerank.} We set the size of word embeddings to 300, the convolution encoder block number to 1, and the kernel size of CNN to 3. The weights are shared between the article and template encoders. The $k$ of k-max pooling is set to 10. L2 weight decay with $\lambda=3\times 10^{-6}$ is performed over all trainable variables. The initial learning rate is 0.001 and multiplied by 0.1 every 10K steps. Dropout between layers is applied.

\noindent\textbf{BiSET.} A two-layer BiLSTM is used as the encoder, and another two-layer LSTM as the decoder. The sizes of word embeddings and LSTM hidden states are both set to 500. We only apply dropout in the LSTM stack with a rate of 0.3. The learning rate is set to 0.001 for the first 50K steps and halved every 10K steps. Beam search with size 5 is applied to search for optimal answers.

\subsection{Evaluation Metrics}
Following previous work \cite{Nallapati2016Abstractive,Zhou2017Selective,cao2018retrieve}, we use the standard F1 scores of ROUGE-1, ROUGE-2 and ROUGE-L \cite{Lin2003Automatic} to evaluate the selected templates and generated summaries, where the official ROUGE script\footnote{The ROUGE evaluation option: -m -n 2 -w 1.2} is applied. We employ the normalized discounted cumulative gain (NDCG) \cite{jarvelin2002cumulated} from information retrieval to evaluate the Fast Rerank module.

\section{Results and Analysis}
In this section, we report our experimental results with thorough analysis and discussions.

\subsection{Performance of Retrieve}\label{sec:retrieve}
The Retrieve module is intended to narrow down the search range for a best template. We evaluated this module by considering three types of templates: (a) \noindent\textbf{Random} means a randomly selected summary from the training corpus; (b) \noindent\textbf{Retrieve-top} is the highest-ranked summary by Retrieve; (c) \noindent\textbf{N-Optimal} means among the $N$ top search results, the template is specified as the summary with largest ROUGE score with gold summary.

\begin{figure}
\centering
\includegraphics[scale=0.24]{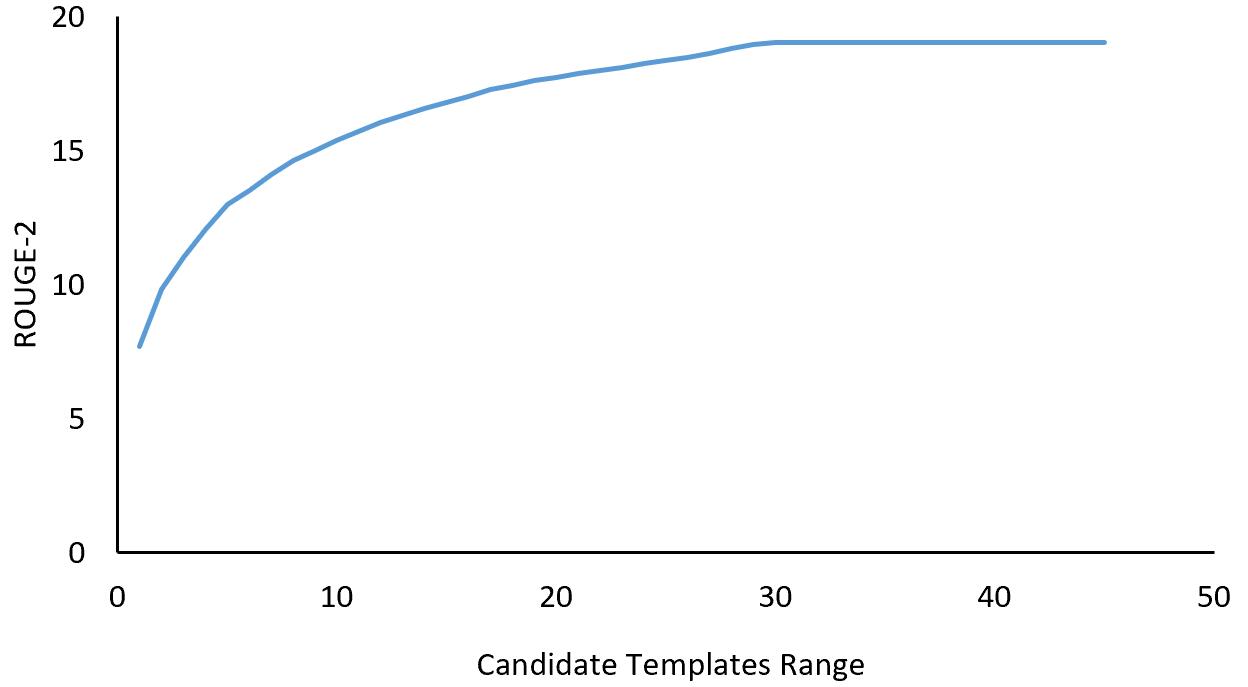}
\caption{Quality of candidate templates under different ranges. }
\label{img:data}
\end{figure}

As the results show in Table \ref{tab:Retrieve}, randomly selected templates are totally irrelevant and unhelpful. When they are replaced by the Retrieve-top templates, the results improve apparently, demonstrating the relatedness of top-ranked summaries to gold summaries.~Furthermore, when the N-Optimal templates are used, additional improvements can be observed as $N$ grows. This trend is also confirmed by Figure \ref{img:data}, in which the ROUGE scores increase before 30 and stabilize afterwards. These results suggest that the ranges given by Retrieve indeed help to find quality templates.

\begin{table}[h]
\small

	\centering
	\begin{tabular}{@{}l|ccc@{}}
		\toprule
        Type     & ROUGE-1   & ROUGE-2  & ROUGE-L   \\
        \hline
		Random       & 2.58 & 0.00 & 2.48 \\
		Retrieve-top      & 23.46 & 7.67 & 20.94 \\
		5\textrm{-}Optimal  & 32.69 & 11.74 & 28.71 \\
        10\textrm{-}Optimal  & 35.90 & 13.32 & 31.42 \\
        15\textrm{-}Optimal  & 37.82 & 16.79 & 34.08 \\
        20\textrm{-}Optimal  & 38.92 & 17.72 & 34.94 \\
        30\textrm{-}Optimal  & 40.49 & 19.01 & 36.10 \\
        \bottomrule
	\end{tabular}
	\caption{Performance of different types of templates.}
	\label{tab:Retrieve}
\end{table}

\subsection{Fast Rerank}\label{sec:fast rerank result}
\begin{figure}
\centering
\includegraphics[scale=0.65]{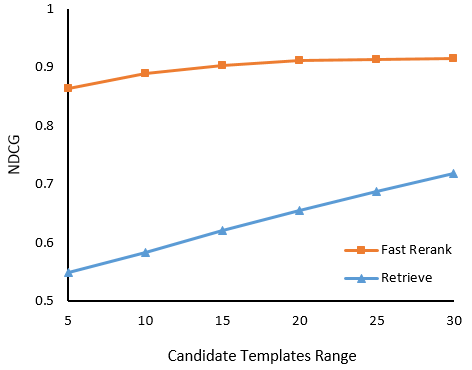}
\caption{Quality of rankings given by Fast Rerank.}
\label{img:ndcg}
\end{figure}

As mentioned before, the role of Fast Rerank is to re-rank the initial search results and return a best template for summarization. To examine the effect of this module, we studied its ranking quality under different ranges as in Section \ref{sec:retrieve}. The original rankings by Retrieve are presented for comparison with the NDCG metric. We regard the ROUGE-2 score of each candidate template with the reference summary as the ground truth.~As shown in Figure \ref{img:ndcg}, Fast Rerank consistently provides enhanced rankings over the original.

\subsection{Interaction Approaches}
In Section \ref{section:interaction_methods}, we also explored three alternative approaches to integrating an article with its template.~The results are shown in Table \ref{tab:Interaction}, from which we can note that none of these approaches help yield satisfactory performance. Even though DCN Attention works impressively in machine reading comprehension, it performs even worse in this task than the simple concatenation. We conjecture the reason is that the DCN Attention attempts to fuse the template information into an article as in machine reading comprehension, rather than  selects key information from the two to form an enhanced article representation.

\begin{table}[h]
\small

	\centering
	\begin{tabular}{@{}l|ccc@{}}
		\toprule
        Interaction method& ROUGE-1   & ROUGE-2  & ROUGE-L  \\
        \hline
        Concatenation  & 32.26 & 15.30 & 30.19 \\
        Concate+multi self-att  & 33.15 & 15.93 & 31.21 \\
        DCN Attention   & 31.53 & 13.77 & 27.96 \\
        Bi-selective layer &39.11 & 19.78 & 36.87 \\
        \bottomrule
	\end{tabular}
	\caption{Results of different interaction approaches.}
\label{tab:Interaction}
\end{table}

\subsection{BiSET}
The overall performance of all the studied models is shown in Table \ref{tab:Rewrite}.~The results show that our model significantly outperforms all the baseline models and sets a new state of the art for abstractive sentence summarization. To evaluate the impact of templates on our model, we also implemented BiSET with two other types of templates: randomly-selected templates and best templates identified by Fast Rank under different ranges.~As shown in Table \ref{tab:Roubustness}, the performance of our model improves constantly with the improvement of template quality (larger ranges lead to better chances for good templates). Even with randomly-selected templates, our model still works with stable performance, demonstrating its robustness.

\begin{table}[h]
\small
	\centering
\scalebox{0.85}{
	\begin{tabular}{@{}p{3.7cm}|p{1.4cm}<{\centering}p{1.4cm}<{\centering}p{1.4cm}<{\centering}@{}}
		\toprule
        Model     & ROUGE-1&ROUGE-2  & ROUGE-L   \\
        \hline
        ABS$^\ddagger$ \cite{Rush2015A}& 29.55 & 11.32 & 26.42 \\
		ABS+$^\ddagger$ \cite{Rush2015A}& 29.78 & 11.89 & 26.97 \\
		RAS-Elman$^\ddagger$ \cite{Chopra2016Abstractive}& 33.78 & 15.97 & 31.15 \\
        Featseq2seq$^\ddagger$ \cite{Nallapati2016Abstractive} & 32.67 & 15.59 & 30.64 \\
		Open-NMT$^\ddagger$ \cite{opennmt}& 34.07 & 16.35 & 31.78 \\
		SEASS$^\ddagger$ \cite{Zhou2017Selective}& 36.15 & 17.54 & 33.63 \\
        S2S+CGU$^\ddagger$ \cite{Lin2018Global}& 36.30 & 18.00 & 33.80 \\
        FTSum$^\ddagger$ \cite{cao2018faithful}& 37.27 & 17.65 & 34.24 \\
        R\textsuperscript{3}Sum$^\ddagger$ \cite{cao2018retrieve}& 37.04 & 19.03 & 34.46 \\
        \hline
        \textbf{BiSET} & \textbf{39.11} & \textbf{19.78} & \textbf{36.87} \\
        \bottomrule
	\end{tabular}
}
	\caption{Performance of all the models, where results marked with $^\ddagger$ are taken from the corresponding papers.}
\label{tab:Rewrite}
\end{table}

\begin{table}[h]
\small
	
	\centering
	\begin{tabular}{@{}l|ccc@{}}
		\toprule
        Template Type & ROUGE-1   & ROUGE-2  & ROUGE-L  \\
        \hline
        Random  & 33.85 & 15.83 & 31.14 \\
        5-rerank  & 37.69 & 18.62 & 34.38 \\
        10-rerank   & 38.34 & 19.35 & 34.97 \\
        20-rerank   & 38.89 & 19.64 & 36.67 \\
        30-rerank   & 39.11 & 19.78 & 36.87 \\

        \bottomrule
	\end{tabular}
	\caption{Performance of BiSET with different types of templates, where \textbf{Random} means randomly-selected templates, and \textbf{N-rerank} denotes the best templates re-ranked by Fast Rerank under range $N$.}
\label{tab:Roubustness}
\end{table}

\subsection{Speed Comparison} Our model is designed for both accuracy and efficiency. Due to the parallelizable nature of CNN, the Fast Rerank module only takes about 30 minutes for training and 3 seconds for inference on the whole test set. The BiSET model takes about 8 hours for training (GPU:GTX 1080), 6 times faster than $R^3Sum$ \cite{cao2018retrieve}\footnote{It takes about 2 days for training.}.

\subsection{Ablation Study}
The purpose of this study is to examine the roles of the bi-directional selective layer and its two gates. Firstly, we removed the selective layer and replaced it with the direct concatenation of an article with its template representation. As the results show in Table \ref{tab:Ablation}, the model performs even worse than some ordinary sequence-to-sequence models in Table \ref{tab:Rewrite}. The reason might be that templates would overwhelm the original article representations and become noise after concatenation. Then, we removed the Template-to-Article (T2A) gate, and as a result the model shows a great decline in performance, indicating the importance of templates in article representations. Finally, when we removed the Article-to-Template (A2T) gate, whose role is to control the weight of T2A in article representations, only a small performance decline is observed. This may suggest that the T2A gate alone can already capture most of the important article information, while A2T plays some supplemental role.

\begin{table}[t]
\small
	
	\centering
	\begin{tabular}{@{}l|ccc@{}}
		\toprule
        Model & ROUGE-1   & ROUGE-2  & ROUGE-L  \\
        \hline
        Concatenation  & 32.26 & 15.30 & 30.19 \\
        BiSET without T2A   & 34.51 & 16.55 & 31.17 \\
        BiSET without A2T  & 39.02 & 19.21 & 36.02 \\
        BiSET(full) & 39.11 & 19.78 & 36.87 \\
        \bottomrule
	\end{tabular}
	\caption{ROUGE F1 scores of ablated models.}
\label{tab:Ablation}
\end{table}

\subsection{Human Evaluation}
We then carried out a human evaluation to evaluate the generated summaries from another perspective. Our evaluators include 8 graduate students and 4 senior undergraduates, while the dataset is 100 randomly-selected articles from the test set.~Each sample in this dataset also includes: 1 reference summary, 5 summaries generated by Open-NMT\footnote{https://github.com/OpenNMT/OpenNMT-py} \cite{opennmt}, $R^3Sum$\footnote{http://www4.comp.polyu.edu.hk/˜cszqcao/} \cite{cao2018retrieve} and BiSET under three settings, respectively, and 3 randomly-selected summaries for trapping. We asked the evaluators to independently rate each summary on a scale of 1 to 5, with respect to its quality in \emph{informativity}, \emph{conciseness}, and \emph{readability}. While collecting the results, we rejected the samples in which more than half evaluators rate the \emph{informativity} of the reference summary below 3. We also rejected the samples in which the \emph{informativity} of a randomly-selected summary is scored higher than 3. Finally, we obtained 43 remaining samples and calculated an average score for each aspect. As the results show in Table \ref{tab:human evaluation}, our model not only performs much better than the baselines, it also shows quite comparable performance with the reference summaries.

\begin{table}[h]
	\centering
	\begin{tabular}{@{}l|ccc@{}}
		\toprule
        Model & Info   & Concise & Read  \\
        \hline
        R\textsuperscript{3}Sum & 3.30 & 3.83 & 3.90 \\
        Open-NMT  & 3.26 & 3.69 & 3.86 \\
        BiSET(random template)  & 3.09 & 3.69 & 3.71 \\
        BiSET(without A2T)& 3.24 & 3.75 & 3.72 \\
        BiSET(best template) & 3.35 & \textbf{3.98} & \textbf{3.93} \\
        Reference& \textbf{3.55} & 3.91 & 3.89 \\
        \bottomrule
	\end{tabular}
	\caption{Results of human evaluation.}
\label{tab:human evaluation}
\end{table}

In Table \ref{tab:Prediction} we present two real examples, which show the templates found by our model are indeed related to the source articles, and with their aid, our model succeeds to keep the main content of the source articles for summarization while discarding unrelated words like `US' and `Olympic Games'.

\begin{table}[t]
\small
	\centering
    \subtable{
	\begin{tabular}{@{}p{1cm}|p{6.0cm}@{}}
		\toprule
        Source&factory orders for manufactured goods rose \#.\# percent in September, the commerce department said here Thursday.\\
        \hline
        Ref&September factory orders up \#.\# percent. \\
         \hline
         Temp&January factory orders in US up \#.\# percent.\\
         \hline
        BiSET&factory orders up \#.\# percent in September.\\
        \bottomrule
	\end{tabular}
    }
        \subtable{
	\begin{tabular}{@{}l|p{6.0cm}@{}}
		\toprule
        Source&some \#.\# billion people worldwide are expected to watch Germany face Costa Rica on television at the opening match of football's World Cup, German public broadcaster zdf said Thursday.\\
        \hline
        Ref&\#.\# billion tv viewers expected for opening World Cup match.\\
         \hline
        Temp&billions around world watch the Olympic Games opening ceremony.\\
        \hline
        BiSET&\#.\# billions around world expected to watch World Cup.\\
        \bottomrule
	\end{tabular}
    }
    \caption{Examples of the generated templates and summaries by our model. `\#' refers to masked numbers.}
	\label{tab:Prediction}
\end{table}

\section{Related Work}
Abstractive sentence summarization, a task analogous to headline generation or sentence compression, aims to generate a brief summary given a short source article. Early studies in this problem mainly focus on statistical or linguistic-rule-based methods, including those based on extractive and compression \cite{jing2000cut,knight2002summarization,clarke2010discourse}, templates \cite{zhou2004template} and statistical machine translation \cite{banko2000headline}.

The advent of large-scale summarization corpora accelerates the development of various neural network methods. \newcite{Rush2015A} first applied an attention-based sequence-to-sequence model for abstractive summarization, which includes a convolutional neural network (CNN) encoder and a feed-forward network decoder. \newcite{Chopra2016Abstractive} replaced the decoder with a recurrent neural network (RNN). \newcite{Nallapati2016Abstractive} further changed the sequence-to-sequence model to a fully RNN-based model. Besides, \newcite{Gu2016Incorporating} found that this task benefits from copying words from the source articles and proposed the CopyNet correspondingly. With a similar purpose, \newcite{Gulcehre2016Pointing} proposed to use a switch gate to control when to copy from the source article and when to generate from the vocabulary. \newcite{Zhou2017Selective} employed a selective gate to filter out unimportant information when encoding.

Some other work attempts to incorporate external knowledge for abstractive summarization. For example, \newcite{Nallapati2016Abstractive} proposed to enrich their encoder with handcrafted features such as named entities and part-of-speech (POS) tags. \newcite{guu2018generating} also attempted to encode human-written sentences to improve neural text generation. Similar to our work, \newcite{cao2018retrieve} proposed to retrieve a related summary from the training set as soft template to assist with the summarization. However, their approach tends to oversimplify the role of the template, by directly concatenating a template after the source article encoding. In contrast, our bi-directional selective mechanism exhibits a novel attempt to selecting key information from the article and the template in a mutual manner, offering greater flexibility in using the template.

\section{Conclusion}
In this paper, we presented a novel \textbf{Bi}-directional \textbf{S}elective \textbf{E}ncoding with \textbf{T}emplate (BiSET) model for abstractive sentence summarization. To counteract the verbosity and insufficiency of training data, we proposed to retrieve high-quality existing summaries as templates to assist with source article representations through an ingenious bi-directional selective layer. The enhanced article representations are expected to contribute towards better summarization eventually. We also developed the corresponding retrieval and re-ranking modules for obtaining quality templates. Extensive evaluations were conducted on a standard benchmark dataset and experimental results show that our model can quickly pick out high-quality templates from the training corpus, laying key foundation for effective article representations and summary generations. The results also show that our model outperforms all the baseline models and sets a new state of the art. An ablation study validates the role of the bi-directional selective layer, and a human evaluation further proves that our model can generate informative, concise, and readable summaries.

\section{Acknowledgement}
The paper was partially supported by the Program for Guangdong Introducing Innovative and Enterpreneurial Teams (No.2017ZT07X355) and the Key R$\&$D Program of Guangdong Province (2019B010120001).

\end{spacing}
\newpage
\bibliography{acl2019}
\bibliographystyle{acl_natbib}

\end{document}